\documentclass[runningheads]{llncs} 
\usepackage{graphicx}
\usepackage{url}

\usepackage[belowskip=-10pt,aboveskip=-1pt]{caption}
\usepackage[belowskip=2pt,aboveskip=2pt]{subcaption}

\begin{document}

\makeatletter
\renewcommand\section{\@startsection{section}{1}{\z@}%
                       {-8\p@ \@plus -4\p@ \@minus -4\p@}%
                       {6\p@ \@plus 4\p@ \@minus 4\p@}%
                       {\normalfont\large\bfseries\boldmath
                        \rightskip=\z@ \@plus 8em\pretolerance=10000 }}
\renewcommand\subsection{\@startsection{subsection}{2}{\z@}%
                       {-8\p@ \@plus -4\p@ \@minus -4\p@}%
                       {6\p@ \@plus 4\p@ \@minus 4\p@}%
                       {\normalfont\normalsize\bfseries\boldmath
                        \rightskip=\z@ \@plus 8em\pretolerance=10000 }}
\renewcommand\subsubsection{\@startsection{subsubsection}{3}{\z@}%
                       {-4\p@ \@plus -4\p@ \@minus -4\p@}%
                       {-1.5em \@plus -0.22em \@minus -0.1em}%
                       {\normalfont\normalsize\bfseries\boldmath}}
\makeatother

\title{Understanding and representing the semantics\\ of large structured documents}
\author{Muhammad Mahbubur Rahman \and
Tim Finin
}
\authorrunning{M. Rahman et al.}
\institute{University of Maryland, Baltimore County, Baltimore MD 21250, USA \\
\email{\{mrahman1,finin\}@umbc.edu}}
\maketitle              % typeset the header of the contribution

\begin{abstract}
Understanding large, structured documents like scholarly articles, requests for proposals or business reports is a complex and difficult task. It involves discovering a document's overall purpose and subject(s), understanding the function and meaning of its sections and subsections, and extracting low level entities and facts about them. In this research, we present a deep learning based document ontology to capture the general purpose semantic structure and domain specific semantic concepts from a large number of academic articles and business documents. The ontology is able to describe different functional parts of a document, which can be used to enhance semantic indexing for a better understanding by human beings and machines. We evaluate our models through extensive experiments on datasets of scholarly articles from \textit{arXiv} and \textit{Request for Proposal} documents. 

\end{abstract}

\section{Introduction}

Understanding the semantic structure of large multi-themed documents is a challenging task because these documents are composed of a variety of functional sections discussing diverse topics. Some documents may have a table of contents, whereas others may not. Even if a table of contents is present, mapping it across the document is not a straightforward process. Section and subsection headers may or may not be present in the table of contents and if they are present, they are often inconsistent across documents even within the same vertical domain. 

Identifying a semantic organization of sections, subsections and sub-subsections of documents across all vertical domains is not the same. For example, a business document has a completely different structure from a user manual. Scholarly research articles from different disciplines, such as computer science and social science, may have different structures. For example, social science articles usually have \textit{methodology} sections whereas computer science articles often have \textit{approach} sections. Semantically these two section types share the same purpose and function, even though their details may be quite different.

Our objective is to develop and use a document ontology to describe different functional parts of academic and business documents. For example, the \textit{introduction} section of a research paper describes the problem statement, scope and context by explaining the significance of the research challenge. The \textit{results} section presents and illustrates research findings with the help of experiments, graphs and tables. Finally, the \textit{conclusion} section typically restates the paper's contribution and the most important ideas that support the main argument of the paper.   

Creating such an ontology involves significant human understanding and analysis of a large number of documents from any vertical domain. It is also requires a common understanding of the structure of information presented in those documents. The common concepts across all documents should be clearly visible to the ontology developers. The developers should also understand the hierarchy of the sections, subsection and sub-subsections of a document. Hence the process to get each relationship among different concepts of a document is time consuming. Moreover, some concepts may be overlooked while analyzing the documents. 

We have developed a deep learning based system to automatically determine ontology concepts and properties from a large number of documents of the same vertical domain. Our approaches are powerful, yet simple, to capture the most important semantic concepts from academic articles and \textit{request for proposal} (RFP) documents. In the course of out this work, we experimented with and evaluated several state of the art technologies, including Variational Autoencoders (VAE) \cite{kingma2013auto}, Convolutional Autoencoders (CAE) \cite{holden2015learning,turchenko2017deep} and LDA \cite{blei2003latent}. 

The ontology can be used for annotating different sections of a document, which helps to understand its semantic structure. It can also be useful for comprehending and modeling types and subtypes of documents. The results can enable the reuse of domain knowledge along with text analysis, content based question answering and semantic document indexing. 

%The rest of the paper is organized as follows. Section 2 describes related work in the document ontology and semantic annotation of a document. Section 3 explains data types and document types we use in our experiments. It also describes \textit{arXiv} and RFP document categories. The fourth section illustrates the technical approaches along with the system architecture. Our experiments and their results are discussed in section 5. Section 6 describes the domain specific semantic concepts. A use case of our ontology is shown in section 7 and section 8 concludes the paper. 

\section{Related Work}

Over the last few years, several ontologies have been developed to describe a document's semantic structure and annotate it with a semantic label. Some of them were designed for academic articles and others deal with other type of documents. 

Ciccarese et al. developed an Ontology of Rhetorical Blocks (ORB) \cite{ciccarese2011ontology} to capture the coarse-grained rhetorical structure of a scientific article. ORB can be used to add semantics to a new article and to annotate an existing article. It divides a scientific article into three components: header, body and tail. The header captures meta-information about the article, such as it's title, authors, affiliations, publishing venue, and abstract. The body adopts the IMRAD structure from \cite{sollaci2004introduction} and contains introduction, methods, results, and discussion. The tail provides additional meta-information about the paper, such as acknowledgments and references. 

Peroni et al. introduced the Semantic Publishing and Referencing (SPAR) Ontologies \cite{peroni2014semantic} to create comprehensive machine-readable RDF meta-data for the entire set of characteristics of a document from semantic publishing. It is used to describe different components of books and journal articles, such as citations and bibliographic records. It has eight ontologies to cover all of the components for the creation of RDF meta-data (DoCO, FaBiO, CiTO, PRO, PSO, C4O, BiRO and PWO). DoCO, the document components ontology \cite{shotton2011doco,constantin2016document}, provides a general-purpose structured vocabulary of document elements to describe both structural and rhetorical document components in RDF. This ontology can be used to annotate and retrieve document components of an academic article based on its structure and content. Examples of DoCO classes are chapter, list, preface, table and figure. DoCO also inherits another two ontologies: Discourse Elements Ontology (Deo) \cite{DEO} and Document Structural Patterns Ontology \cite{PatternOntology}.

Shotton et al. developed the Deo ontology to study different corpora of scientific literature on different topics and publishers. It presents structured vocabulary for rhetorical elements within an academic document. The major classes of Deo are introduction, background, motivation, model, related work, methods, results, conclusion, and acknowledgements. This ontology is very intriguing and relevant to our semantic annotation. 

Monti et al. developed a system to reconstruct an electronic medical document with semantic annotation \cite{monti2016semantic}. They divided the process into three steps. In the first, they classified documents in one of the categories specified in the Consolidated CDA (C-CDA) standard \cite{dolinhl7}, using PDFBox \cite{wiki:pdfbox} to extract text from CDA standard medical documents. Later, they split the document into paragraphs using the typographical features available in the PDF file. Finally, they identified key concepts from the document and mapped them to the most appropriate medical ontology. However, the paper lacks technical detail and an analysis of the results.

A Contextual Long short-term memory (CLSTM) \cite{hochreiter1997long} was used by Ghosh et al. for sentence topic prediction \cite{ghosh2016contextual}. Lopyrev et al. trained an encoder-decoder RNN with LSTM for generating news headlines using the texts of news articles from the Gigaword dataset \cite{lopyrev2015generating}. Srivastava et al. introduced a type of Deep Boltzmann Machine (DBM) for extracting distributed semantic representations from a large unstructured collection of documents \cite{srivastava2013modeling}. They used the Over-Replicated $Softmax$ model for document retrieval and classification.

Tuarob et al. described an algorithm to automatically build a semantic hierarchical structure of sections for a scholarly paper \cite{tuarob2015hybrid}. They defined a section as the pair of the section header and its textual content. They employed a rule-based approach to recognize sections from scholarly articles and applied a simple set of heuristics that built a hierarchy of sections from the extracted section headers.  

Most of the ontologies mentioned above are developed either manually or do not provide any technical details. 
Our ontology is an enhancement of Deo\cite{DEO}, developed using deep learning and embedding vector clustering to choose classes and the properties based on more than $1$ million academic articles and a few hundred thousand business documents. Our approach is able to capture semantic meaning of different functional parts of a document by semantic annotation and semantic concept extraction, as described in our earlier research \cite{rahman2017deep}. 

\section{Data Type and Document Category}

In this research, we focus on extracting information from PDF documents. The motivation for focusing on PDF documents is the popularity and portability of PDF over different types of devices and operating systems. But automatic post-processing of a PDF document is not an easy task, since the objective of PDF rendering tools is not to support post-processing, but rather better visualization of the content. The rendering tools allow numerous equally valid ways of producing the same visual result and therefore no structure can reliably be derived from how the text operators are used. For experimental purposes, we choose PDF documents from academic articles and business documents, such as \textit{arXiv} and RFP domains. We use a dataset which contains $1,121,363$ \textit{arXiv} articles during or before 2016 released by Rahman et al.\cite{rahman2017understanding}.

\section{System Architecture and Technical Approach}
\label{sec:approach}

This section describes the overall work flow of our system and the approach used for each of its parts.

\begin{figure}[!t]
\begin{center}
\includegraphics[height=1.8in, width=4.0in]{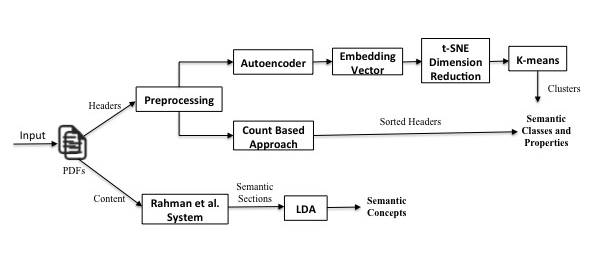}
\caption{Overall work flow of our system\label{fig:pipe_line_ontology}}
\end{center}
\end{figure}

\subsection{System Architecture}

The pipeline of our system is shown in Figure \ref{fig:pipe_line_ontology}. The top-level, subsection and sub-subsection headers are retrieved from all the \textit{arXiv} articles released by Rahman et al. \cite{rahman2017understanding}. After preprocessing, the headers are passed into Autoencoder. The embedding vector is dumped from the Autoencoder, which is passed through a t-SNE \cite{maaten2008visualizing} dimensionality reduction. A k-mean \cite{hartigan1979algorithm} clustering algorithm is then applied on the reduced embedding vector. After preprocessing, a count based approach is also applied to retrieve all of the unique headers based on count. We also use our previous system \cite{rahman2017deep} to get semantic sections, which are passed through LDA topic models to get domain specific semantic terms or concepts for each of the individual sections. 

\subsection{Technical Approach}

In order to design a document ontology, we created a list of classes and properties by following the count-based and cluster-based approaches. In the count-based approach, we first took all section headers, including \textit{top-level}, \textit{subsection} and \textit{sub-subsection} which are basically headers from the table of contents of all \textit{arXiv} articles. Then we removed numbers and dots from the beginning of each header, and generated the frequency for each header and sorted them. Based on a frequency threshold, we considered the section headers, which might be a class or concept for our ontology. 

For the cluster based approach, we generated all section headers from the table of contents of all \textit{arXiv} articles and developed a Variational Autoencoder and Convolutional Autoencoder to represent each of the section headers in a sentence level embedding, which is termed "header embedding" in our system. We applied Autoencoder to learn the header embedding in an unsupervised fashion, in order to produce good quality clusters.  We then dumped the embedding vector from the bottleneck layer. Since this vector has high dimensionality and clustering high-dimensioned data often does not work well, we applied the t-SNE dimensionality reduction technique to reduce the dimensions of the embedding vector to just two dimensions. After dimensionality reduction, we used k-means clustering on the embedding vector to cluster the header embedding into semantically meaningful groups. We analyzed all clusters and all section headers from the count-based approach and came up with the classes or general purpose semantic concepts to design our document ontology. 

\begin{table}[!t]
\centering
\caption{Classes for Ontology}
\label{ClassesforOntology}
\resizebox{\columnwidth}{!} {
\begin{tabular}{|c|l|}
\hline
\textbf{Document Type}    & \multicolumn{1}{c|}{\textbf{Classes/Concepts}}    \\ \hline
\textbf{Academic Article} & \begin{tabular}[c]{@{}l@{}}Introduction, Conclusion, Discussion, References, Acknowledgments, Results, Abstract, \\ Appendix, Related Work, Experiments, Methodology, Proof of Theorem, Evaluation, \\ Future Work, Datasets, Contribution, Background, Implementation, Approach, Preliminary\end{tabular} \\ \hline
\textbf{RFP}              & \begin{tabular}[c]{@{}l@{}}Introduction, Requirement, General Information, Conclusion, Statement of Work, \\ Contract Administration, Appendix, Background, Deliverable, Contract Clauses\end{tabular}                                                                                                     \\ \hline
\end{tabular}}
\end{table}

We also applied a similar approach for section headers from RFP documents. To understand the sections of an RFP, we read \cite{SectionsofanRFP} and discussed with experts from RedShred \cite{RedshRed}. Table \ref{ClassesforOntology} shows classes from \textit{arXiv} articles and RFPs, which were used to design a simple document ontology. The detailed descriptions are given below.

\begin{figure}[!t]
\begin{center}
\includegraphics[height=1.5in, width=4.0in]{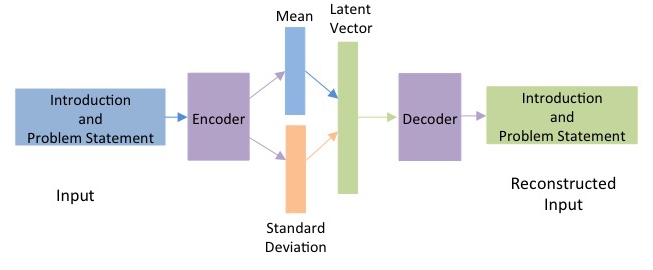}
\caption{Variational Autoencoder for Ontology Class Selection\label{fig:vae_for_ontology}}
\end{center}
\end{figure}

% http://kvfrans.com/variational-autoencoders-explained/
% https://blog.keras.io/building-autoencoders-in-keras.html

\subsubsection{Variational Autoencoder}

A variational autoencoder is a type of autoencoder that learns latent variable models \cite{eisenstein2010latent} for the input data. Instead of learning an arbitrary function, the autoencoder learns the parameters of a probability distribution of the input data. The encoder turns the input data into two parameters in a latent space, which are noted as \begin{math}\bar{z} \end{math} and \begin{math} z\log\sigma \end{math}. Then, randomly, a similar data point, $z$ is selected from the latent normal distribution using Equation \ref{eq:vae}. 
\begin{equation}
\label{eq:vae}
  z= \bar{z} + e^{z\log\sigma}*\epsilon
\end{equation}
A final decoder maps these latent space points back to the original input data. The architecture of our VAE is given in Figure \ref{fig:vae_for_ontology}.

\subsubsection{Convolutional Autoencoder}

\begin{figure}[!t]
\begin{center}
\includegraphics[height=1.6in, width=4.2in]{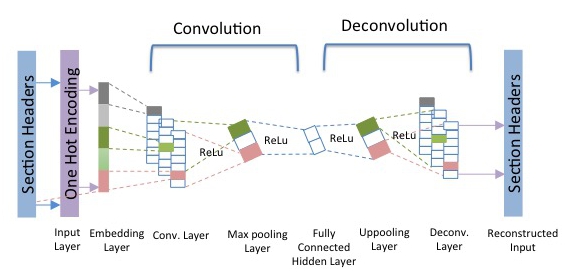}
\caption{Convolutional Autoencoder for Ontology Class Selection\label{fig:Convolutional_Autoencoder}}
\end{center}
\end{figure}

A convolutional autoencoder is an autoencoder that employs a convolutional network to learn the parameters in an unsupervised way. Since our input is text, we use a Conv1D layer for both convolutional and deconvolutional parts of the network. The input text is converted into a \textit{one-hot} encoding, which is passed into the embedding layer. Before encoding, we have Conv1D and MaxPooling1D layers with a \textit{ReLu} activation function. The decoder starts with the deconvolution followed by an UpSampling1D layer. At the end, the decoder reproduces the original text. The architecture of the CAE is given in Figure \ref{fig:Convolutional_Autoencoder}. 

\subsubsection{Document Ontology}
\begin{figure}[!t]
\begin{center}
\includegraphics[height=1.8in, width=4.0in]{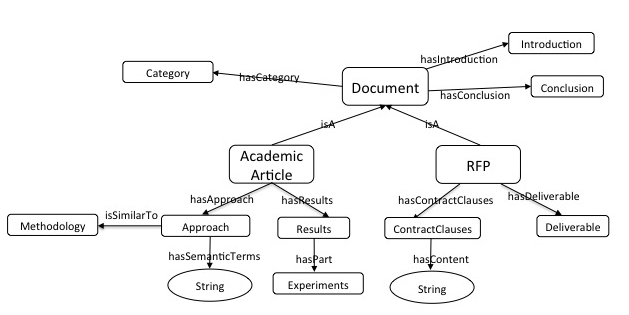}
\caption{The upper level of our document ontology, with rectangles representing classes and ovals properties. Additional classes are mentioned in Table \ref{ClassesforOntology}.}
\label{fig:document_ontology_design_diagram}
\end{center}
\end{figure}

After getting the classes from an analysis of the count- and cluster-based approaches, we designed an ontology for our input documents. The classes represent general purpose semantic concepts in our ontology. We also analyzed cluster visualization to get properties and relations among classes.  Detailed results are included in section \ref{sec:experiments}.  Figure \ref{fig:document_ontology_design_diagram} shows our simple document ontology.

The document ontology includes classes that describe concepts induced from both \textit{arXiv} academic articles and RFP documents. The top level ``Document'' class has two subclasses: ``Academic Article'' and ``RFP''. The Document class has ``Category'' that describes the type of document, such as Computer Science, Mathematics, Social Science, Networking, Biomedical and Software articles/RFPs. Both Academic articles and RFPs have contents, which are sections. These sections are the classes of different semantic concepts in a document. Both Academic articles and RFPs share some concepts, such as ``Introduction'', ``Conclusion'' and ``Background''. They also have their own concepts. For example, ``Approach'' and ``Results'' are available in Academic Articles whereas RFP has ``ContractClauses'' and ``Deliverable'' concepts/classes. Due to space constraint, the classes are shown in Table \ref{ClassesforOntology}.

Each of the classes/concepts has two properties ``SemanticTerms'' and ``Content'' which are represented by the relationships ``hasSemanticTerms'' and ``hasContent''. The data types for these two properties are String. The ``hasSemanticTerms'' property captures semantic topics applying Latent Dirichlet Allocation (LDA) to each section. Some concepts may have part, which is represented by a relationship ``hasPart''. For example, a concept ``Results'' has another subconcept ``Experiments''.  Some of the concepts may be similar to another concepts, which is shown by a relationship ``isSimilarTo''. For example ``Approach'' and ``Methodology'' are two similar concepts. 

\subsubsection{Semantic Concepts using LDA}
We used latent Dirichlet allocation (LDA) \cite{blei2003latent} to find domain-specific semantic concepts from a section. LDA is a generative topic model that is used to understand the hidden structure of a collection of documents. In an LDA model, each document has a mixture of various topics with a probability distribution. Again, each topic is a distribution of words. Using Gensim \cite{vrehuuvrek2011gensim,rehurek_lrec}, we trained an LDA topic model on a set of semantically divided sections. The model is used to predict the topics for any text section. A few terms that have the highest probability values of the predicted topics, are used as domain specific semantic concepts, or terms, for a given section. These semantic concepts are also used as property values in the document ontology. 

\section{Experiments and Results}
\label{sec:experiments}

In this section, we discuss the experimental setup followed by the detailed procedures. We also describe the results and the findings of each experiment and illustrate the results using comparative analysis.

\subsection{Dataset}

Using the dataset released by Rahman and Finin \cite{rahman2017understanding}, we retrieved section headers from table of contents of all \texttt{arXiv} articles and applied some heuristics to remove unwanted text from the headers (e.g.,  numbers and dots) and downcased the text. The total number of unique section headers in our collection was $3,364,668$ for all categories of \textit{arXiv} articles. We used these section headers to get classes or concepts for ontology design as explained in section \ref{sec:approach}. We also retrieved section headers from only Computer Science articles. After applying a similar approach, we found $666,877$ unique section headers from Computer Science articles. The experiments and results for all categories, as well as Computer Science, are described below.

\subsection{Experiments on the arXiv Dataset}
As described in section \ref{sec:approach}, we trained a VAE model to learn the header embedding for ontology design. We clustered the header embedding matrix into semantically meaningful groups and identified different classes for ontology. The VAE was trained with different configurations and hyperparameters to achieve the best results. We experimented with different input lengths, such as $10$, $15$ and $20$ word length section headers. All section headers were converted into a multi-level \textit{one-hot} vector. 

We used $100$ embedding dimensions, $100$ hidden layers and $1.0$ \begin{math} \epsilon \end{math} to learn latent variables. The \textit{one-hot} vector was the input to the network, which was followed by an embedding layer with $ReLu$ activation function. Then we had a dense layer to capture input features in a latent space. The model parameters were trained using two loss functions, which were a reconstruction loss to force the decoded output to match with the initial inputs, and a KL divergence between the learned latent and prior distributions. The decoder was used with a $sigmoid$ activation function and the model was compiled with an $rmsprop$ optimizer and KL divergence loss function. 

\begin{figure*}[!t]
\centering
\captionsetup[subfigure]{justification=centering}
\begin{subfigure}[b]{0.50\textwidth}
  \centering
	\includegraphics[height=1.3in, width=2.4in]{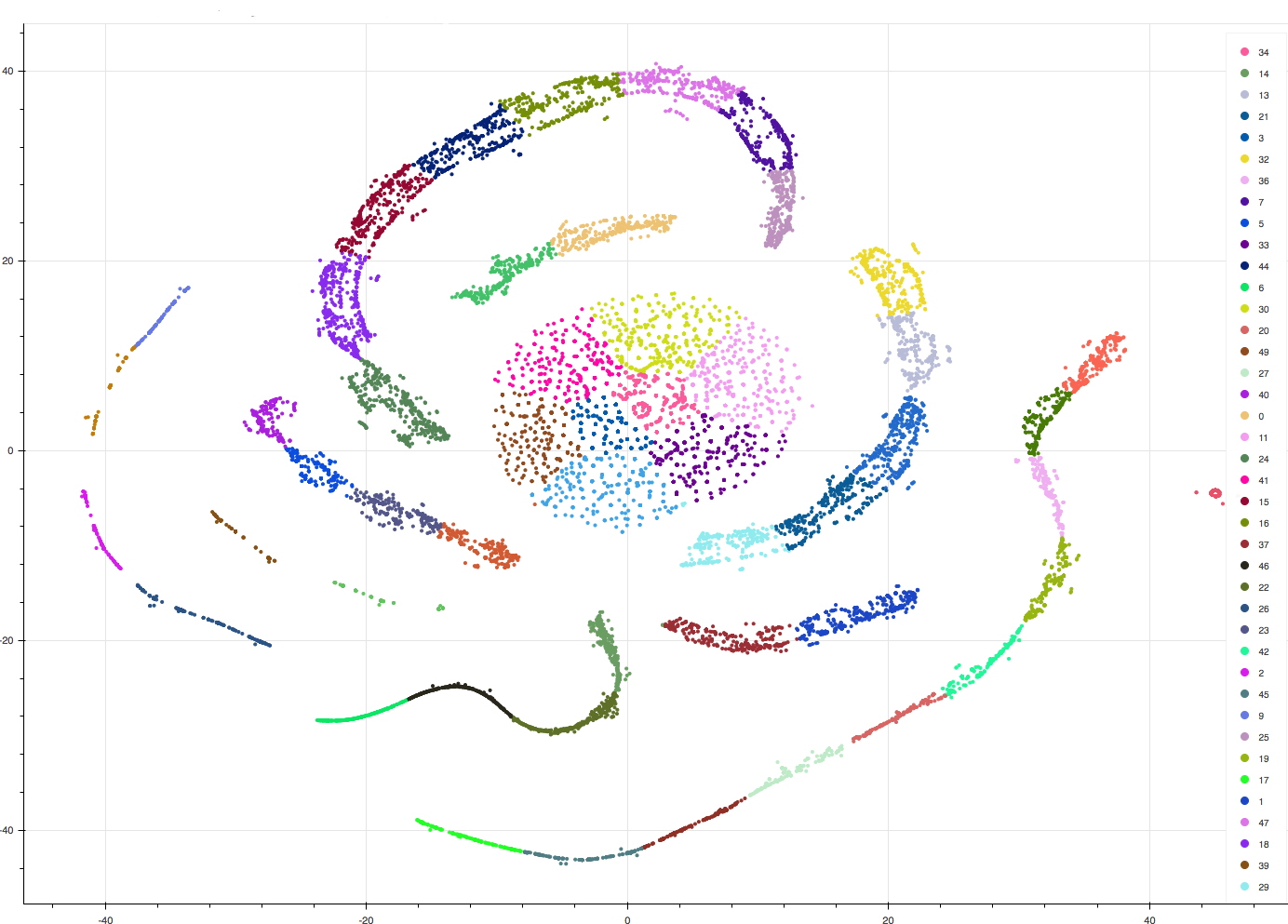}
	\caption{VAE for arXiv: Input Length 15 \label{ontology_encoder_vae_15_clusters_50_encoded_layer}}
\end{subfigure}%
\begin{subfigure}[b]{0.50\textwidth}
  \centering
	\includegraphics[height=1.3in, width=2.4in]{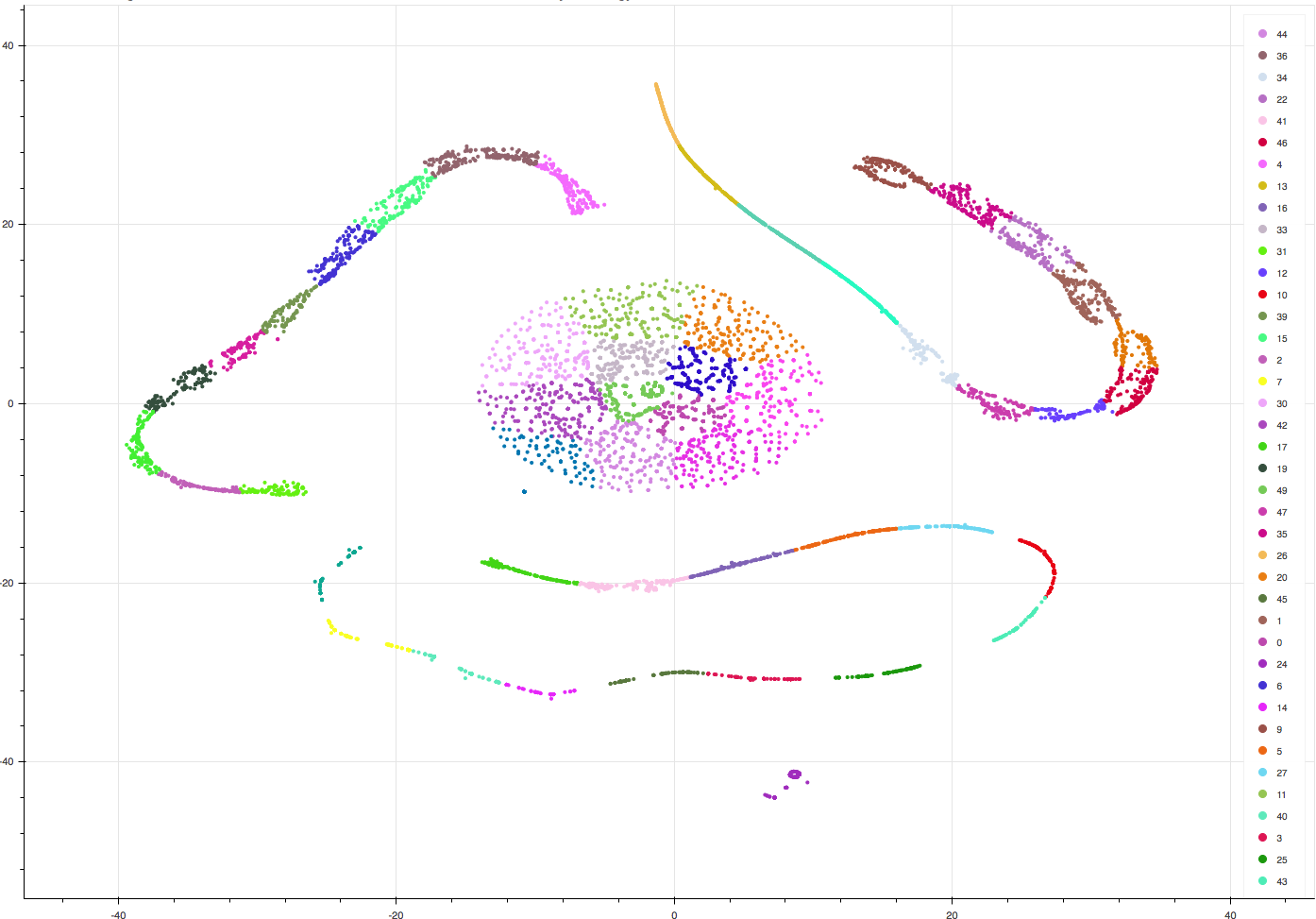}
	\caption{VAE for arXiv: Input Length 20 \label{ontology_encoder_vae_20_clusters_50_encoded_layer}}
\end{subfigure} % 

\begin{subfigure}[b]{0.50\textwidth}
  \centering
	\includegraphics[height=1.3in, width=2.4in]{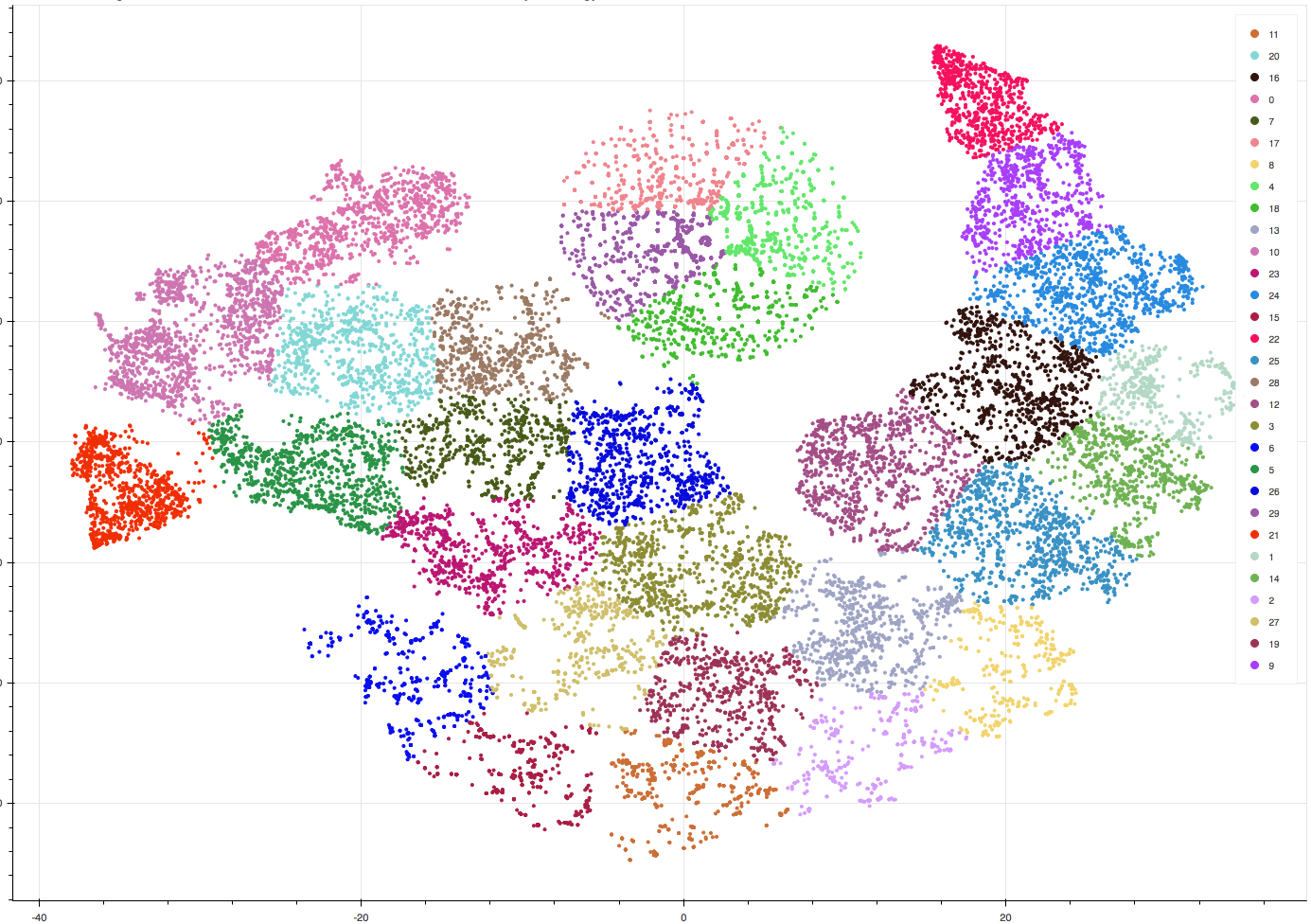}
\caption{CAE for arXiv: Input Length 15 \label{conv_autoencoder_ontology_encoder_clusters_30_layer_layer}}

\end{subfigure}% 
\begin{subfigure}[b]{0.50\textwidth}
  \centering
	\includegraphics[height=1.3in, width=2.4in]{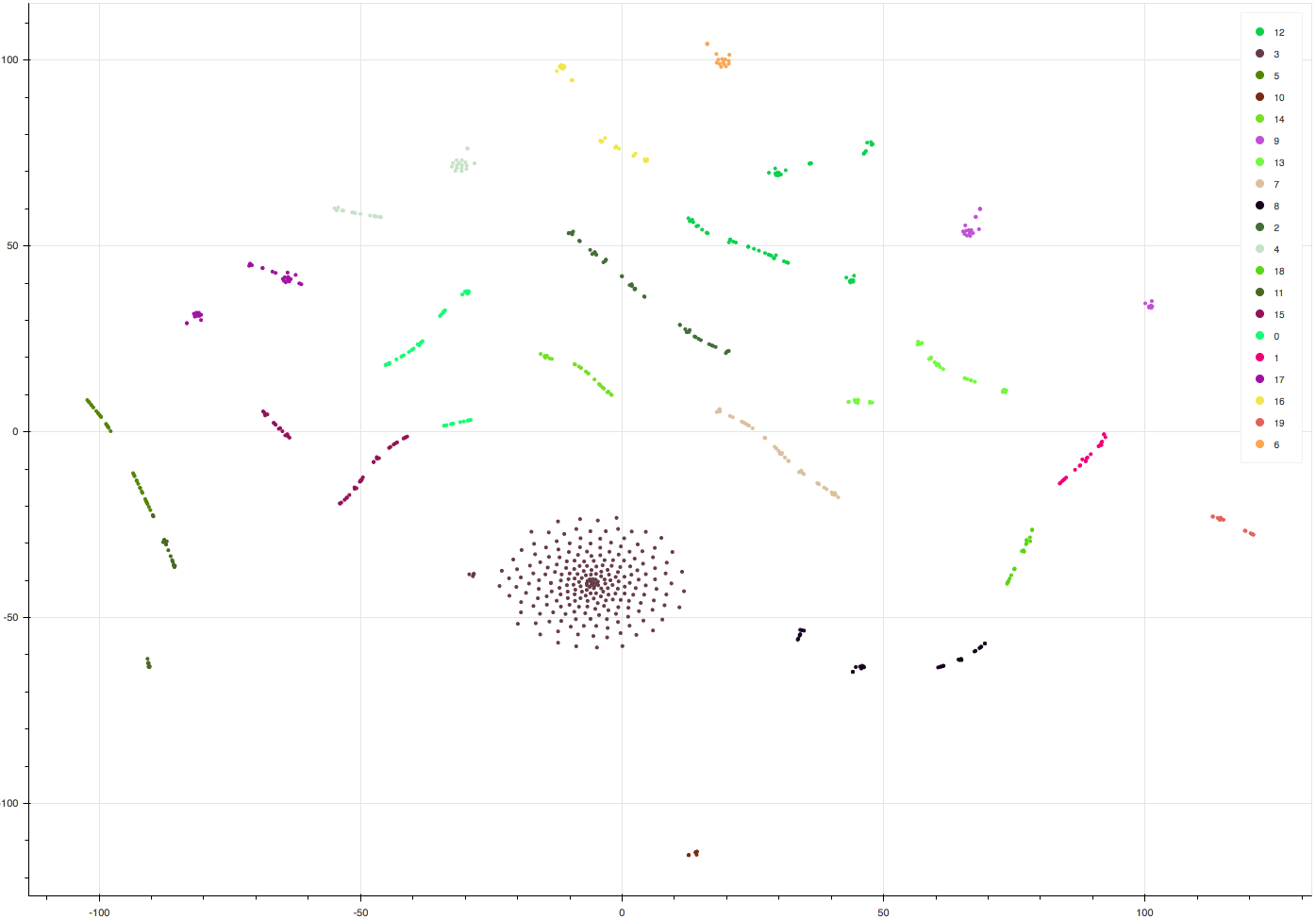}
\caption{VAE for RFP \label{ontology_encoder_vae_15_clusters_20_encoded_layer_rfp}}
\end{subfigure}
\caption{t-SNE Visualization of Embedding Matrix Clusters \label{fig:T-SNEVisualizationofEmbeddingVectorsforVAE}}

\end{figure*}

We also trained a CAE model as described earlier and experimented with different hyperparameters. The model was trained with a $sigmoid$ activation function, $binary\_crossentropy$ loss function, and $adam$ optimizer. We also dumped and clustered the embedding matrix to get semantically meaningful header groups.  

\subsubsection{Results and Evaluation}

Both the VAE and CAE models were trained in an unsupervised way to capture the semantic meaning of each section header. The outputs of the bottleneck layer were dumped and clustered after \textit{t-SNE} dimensionality reduction. Figure \ref{ontology_encoder_vae_15_clusters_50_encoded_layer} shows the visualization of \textit{k-means} clustering with $k=50$ and $input length=15$ for VAE embedding after \textit{t-SNE} dimensionality reduction. Similar visualization with $input length=20$ is shown in Figure \ref{ontology_encoder_vae_20_clusters_50_encoded_layer}. After analyzing both the visualizations, we observed that VAE models learned very well and were able to capture similar section headers together. We noticed that semantically similar section headers were plotted nearby. We also realized that semantically similar section headers were constructed gradually from one concept to another. For example, we detected a pattern in the graph where a sequence of concepts from ``methods'' gradually moved to ``data construction'', ``results'', ``discussion'', ``remarks'' and ``conclusion''. From this analysis, we could infer that VAE learned concepts over section headers in a semantic pattern. From the analysis and visualization of VAE, we found that the VAE models were capable of learning a manifold in the section header embeddings. This manifold can be used for computing semantically similar concepts in our ontology. 

Figure \ref{conv_autoencoder_ontology_encoder_clusters_30_layer_layer} shows the clustering visualization of the embedding matrix generated by the CAE. After analyzing the cluster visualization, we observed that CAE models were not as good as VAE models. We also noticed that CAE models were not capable of learning a manifold in our dataset.

\begin{figure*}[!t]
\centering
\captionsetup[subfigure]{justification=centering}
\begin{subfigure}[b]{0.50\textwidth}
  \centering
	\includegraphics[height=1.0in, width=2.5in]{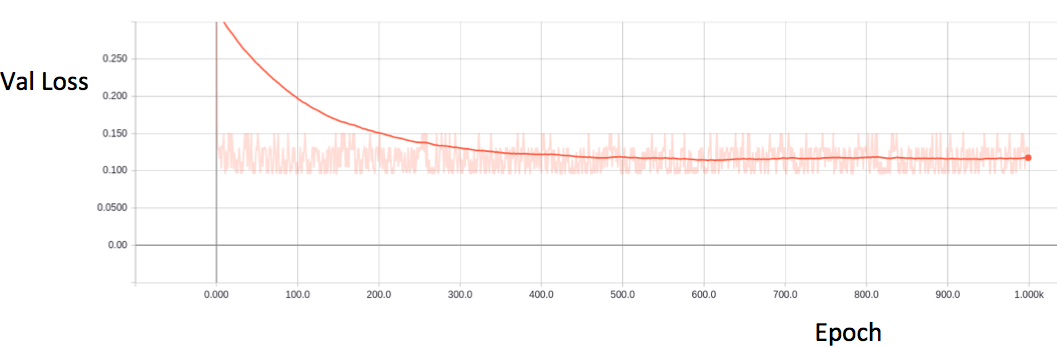}
\caption{For VAE \label{vae_20_val_loss}}

\end{subfigure}% 
\begin{subfigure}[b]{0.50\textwidth}
  \centering
	\includegraphics[height=1.0in, width=2.5in]{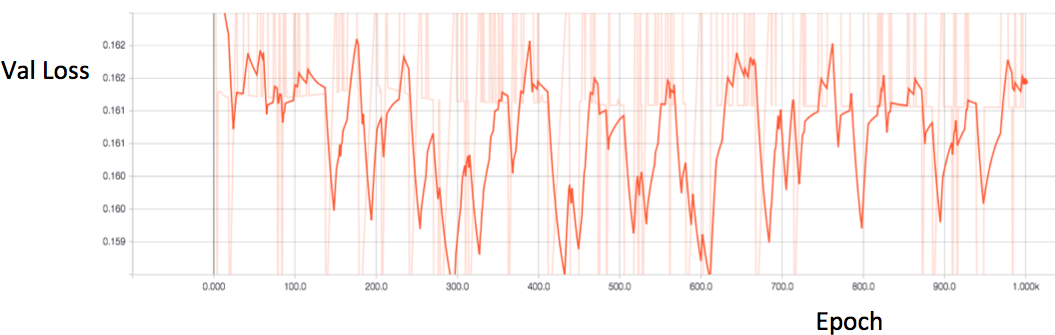}
\caption{For CAE \label{cae_val_loss}}
\end{subfigure}
\caption{Validation loss \label{fig:validation_Loss_vae_cae}}

\end{figure*}

% more evaluation using loss and accuracies. and graphs.....
We achieved the best validation loss of $0.0947$ for the VAE model and $0.1574$ for the CAE model. Figures \ref{vae_20_val_loss} and \ref{cae_val_loss} show the validation losses over number of epochs for the VAE and CAE models, respectively. After analyzing these losses, we observed that VAE loss was steady after $600$ epochs but that the CAE loss was oscillating after $20$ epochs. This suggests that the VAE model performs better than the CAE model for our dataset. Since we achieved a better performance for VAE architecture on our dataset, we also trained VAE models for section headers from Computer Science articles, achieving performance metrics similar to those for all \textit{arXiv} articles.

\subsection{Experiment on RFP Dataset}
%http://delivery.acm.org/10.1145/2830000/2820918/a18-holden.pdf?ip=130.85.233.31&id=2820918&acc=ACTIVE%20SERVICE&key=5F8E7AA76238C9EB%2EE2B546BDBAFC5578%2E4D4702B0C3E38B35%2E4D4702B0C3E38B35&__acm__=1522100952_f0f5c76b30a7c442407ff9ea3f561c87

We leveraged our existing collaboration with RedShred \cite{RedshRed} to get a wide range of RFPs for our experiments. We trained VAE models for section headers from RFP documents. Due to a fewer number of section headers collected from RFP documents, we obtained different patterns, where most of the section headers were scattered all over the embedding space. Figure \ref{ontology_encoder_vae_15_clusters_20_encoded_layer_rfp} shows the embedding visualization  for RFP documents. It is interesting to notice that the VAE models for the RFP dataset are also capable of learning manifold, which can be used for calculating similar concepts.

\begin{table}[!t]
\centering
\caption{Comparative analysis of LDA models for semantic concepts}
\label{table_ComparativeanalysisofLDAmodelsforsemanticconcepts}
\resizebox{\columnwidth}{!} {
\begin{tabular}{|l|l|l|l|}
\hline
\multicolumn{1}{|c|}{\textbf{arXiv Category}}                                                                    & \multicolumn{1}{c|}{\textbf{Word based LDA}}                                                       & \multicolumn{1}{c|}{\textbf{Bigram based LDA}}                                                                                   & \multicolumn{1}{c|}{\textbf{Phrase based LDA}}                                                                      \\ \hline
\textbf{\begin{tabular}[c]{@{}l@{}}Mathematics - Algebraic Topology,\\ Mathematics - Combinatorics\end{tabular}} & \begin{tabular}[c]{@{}l@{}}algebra, lie, maps, \\ element and metric\end{tabular}                  & \begin{tabular}[c]{@{}l@{}}half plane, complex plane, \\ real axis, rational functions\\ and unit disk\end{tabular}              & \begin{tabular}[c]{@{}l@{}}recent, paper is, theoretical,\\ framework, and developed\end{tabular}                   \\ \hline
\textbf{Nuclear Theory}                                                                                          & \begin{tabular}[c]{@{}l@{}}phase, spin, magnetic, \\ particle and momentum\end{tabular}            & \begin{tabular}[c]{@{}l@{}}form factor, matrix elements, \\ heavy ion,  transverse \\ momentum and u'energy loss\end{tabular}    & \begin{tabular}[c]{@{}l@{}}scattering, quark, momentum,\\ neutron move and gcd\end{tabular}                         \\ \hline
\textbf{\begin{tabular}[c]{@{}l@{}}Computer Science - Computer Vision \\ and Pattern Recognition\end{tabular}}   & \begin{tabular}[c]{@{}l@{}}network, performance, \\ error, channel and average\end{tabular}        & \begin{tabular}[c]{@{}l@{}}neural networks, machine \\ learning, loss function,  \\ training data and deep learning\end{tabular} & \begin{tabular}[c]{@{}l@{}}learning, deep, layers, image\\ and machine learning\end{tabular}                        \\ \hline
\textbf{Mathematical Physics}                                                                                    & \begin{tabular}[c]{@{}l@{}}quantum, entropy, \\ asymptotic, boundary and \\ classical\end{tabular} & \begin{tabular}[c]{@{}l@{}}dx dx, initial data, unique \\ solution, positive constant\\ and uniformly bounded\end{tabular}       & \begin{tabular}[c]{@{}l@{}}stochastic, the process of, \\ convergence rate, diffusion \\ rate and walk\end{tabular} \\ \hline
\textbf{\begin{tabular}[c]{@{}l@{}}Astrophysics - Solar and\\ Stellar Astrophysics\end{tabular}}                 & \begin{tabular}[c]{@{}l@{}}stars, emission, \\ gas, stellar and velocity\end{tabular}              & \begin{tabular}[c]{@{}l@{}}active region, flux rope, \\ magnetic reconnection,\\ model set and solar cycle\end{tabular}          & \begin{tabular}[c]{@{}l@{}}magnetic ray, the magnetic, \\ plasma, shock and rays\end{tabular}                       \\ \hline
\end{tabular}}
\end{table}

\section{Domain Specific Semantic Concepts using LDA}

As described in section \ref{sec:approach}, we trained an LDA model using the divided sections generated from \textit{arXiv} articles by the system developed in our earlier work \cite{rahman2017deep}. The total number of training and test sections for the LDA were $128,505$ and $11,633$, respectively. We applied different experimental approaches using word, phrase and bigram dictionaries. The word-based dictionary contains only unigram terms whereas the bigram dictionary has only bigram terms. The phrase-based dictionary contains combination of unigram, bigram and trigram terms. All three dictionaries were developed from the training dataset by ignoring terms that appeared in less than $20$ sections or in more than $10\%$ of the sections of the whole training dataset. The final dictionary size, after filtering, was $100,000$. Different LDA models were trained based on various number of topics and passes. We ran the trained model to identify a topic for any section, which was used to retrieve top terms with the highest probability. The terms with the highest probability were used as a domain specific semantic concepts for a section. 

For the evaluation, we loaded the trained LDA models and generated domain specific semantic concepts from $100$ \textit{arXiv} abstracts, where we knew the categories of the articles. We analyzed their categories and semantic terms. We noticed a very interesting correlation between the \textit{arXiv} category and the semantic terms from LDA topic models, finding that most of the top semantic terms were strongly co-related to their original \textit{arXiv} categories. A comparative analysis is shown in Table \ref{table_ComparativeanalysisofLDAmodelsforsemanticconcepts}. After analysis of the results, we noticed that a bigram LDA model was more meaningful than either of the other two models. 

%\section{Examples}
%Is it possible to generate some Sparql query? 

% \section{Semantic Annotation}

\section{Conclusion}

Semantic annotation can be described as a technique of enhancing a document with automatic annotations that provide a human-understandable way to represent a document's meaning. It also describes the document in such a way that the document is understandable to a machine. Using our developed ontology, we built a system to annotate a PDF document with human understandable semantic concepts from the ontology. The system, along with the research components and ontology, will be available soon\footnote{These will be available in summer 2018.}. In this research, We have presented Variational and Convolutional Autoencoders which capture general purpose semantic structure and different LDA models for domain specific semantic concept extraction from low level representation of large documents. Our approaches are able to detect semantic concepts and properties from a large number of academic and business documents in an unsupervised way. 

\section*{Acknowledgment}

The work was partially supported by National Science Foundation grant 1549697 and a gifts from IBM and Northrop Grumman.


\bibliographystyle{splncs04}
\bibliography{references}

\end{document}